\documentclass{bmvc2k}

\usepackage{times}
\usepackage{epsfig}
\usepackage{graphicx}
\usepackage{amsmath}
\usepackage{amssymb}

\usepackage{textcomp}
\usepackage{nicefrac}
\usepackage{xcolor}
\usepackage{makecell}

\usepackage{wrapfig}

\usepackage{multicol}
\usepackage{tabularx,booktabs}
\usepackage{lipsum}
\usepackage{gensymb}
\usepackage{multirow}
\usepackage{multirow}
\usepackage{arydshln}
\usepackage{enumitem}
\usepackage{pifont}
\newcommand{\xmark}{\ding{55}}%
\newcolumntype{M}[1]{>{\centering\arraybackslash}m{#1}}
\graphicspath{{images/}}
\usepackage{algorithm}
\usepackage{algorithmicx}
\usepackage{mathtools}
\usepackage{amsmath}
\usepackage{amsfonts}
\usepackage{siunitx} 
\usepackage{textcomp}
\usepackage{algpseudocode}

\usepackage{comment}
\usepackage{verbatim}

\title{Unsupervised Spatio-temporal Latent Feature Clustering for Multiple-object Tracking and Segmentation}

\addauthor{Abubakar Siddique}{abubakar.siddique@marquette.edu}{1}
\addauthor{Reza Jalil Mozhdehi}{reza.jalilmozhdehi@marquette.edu}{2}
\addauthor{Henry Medeiros}{henry.medeiros@marquette.edu}{2}

\addinstitution{Opus College of Engineering \\
Marquette University\\
 Milwaukee, USA
}

\runninghead{Siddique et al.}{Spatio-temporal Latent Feature Clustering for MOTS}

\begin{document}

\maketitle

\begin{abstract}
Assigning consistent temporal identifiers to multiple moving objects in a video sequence is a challenging problem. A solution to that problem would have immediate ramifications in multiple object tracking and segmentation problems. We propose a strategy that treats the temporal identification task as a spatio-temporal clustering problem. We propose an unsupervised learning approach using a convolutional and fully connected autoencoder, which we call deep heterogeneous autoencoder, to learn discriminative features from segmentation masks and detection bounding boxes. We extract masks and their corresponding bounding boxes from a pretrained instance segmentation network and train the autoencoders jointly using task-dependent uncertainty weights to generate common latent features. We then construct constraints graphs that encourage associations among objects that satisfy a set of known temporal conditions. The feature vectors and the constraints graphs are then provided to the kmeans clustering algorithm to separate the corresponding data points in the latent space. We evaluate the performance of our method using challenging synthetic and real-world multiple-object video datasets. Our results show that our technique outperforms several state-of-the-art methods. Code and models are available at \url{https://bitbucket.org/Siddiquemu/usc_mots}.
\end{abstract}
\section{Introduction}
\label{sec:intro}
The goal of Multiple Object Tracking and Segmentation (MOTS) algorithms is to establish temporally consistent associations among segmentation masks of multiple objects observed at different frames of a video sequence. To accomplish that goal, most state-of-the-art MOTS methods \cite{MOTSNet, MOTS} employ supervised learning approaches to generate discriminative embeddings 
and then apply feature association algorithms based on sophisticated target behavior models \cite{DBLP:tracking_by_animation, MOTSFusion, GMPHD2020online}. This paper proposes a novel perspective on the problem of temporal association of segmentation masks based on spatio-temporal clustering strategies. 

Subspace clustering algorithms applied to sequential data can separate sequences of similar data points into disjoint groups. State-of-the-art subspace clustering methods have shown promising performance on single object patches \cite{elhamifar2013sparse}, video sequences \cite{facr_cluster_online_video}, and face tracking datasets \cite{video_face_track}. For video sequences containing multiple objects, subspace clustering can be used as a data association strategy to assign a unique temporal identifier to each object. However, due to variations in the data distribution caused by changes in the appearance of the objects and by misdetections, occlusions, and fast motions, subspace clustering in video segments using only location \cite{spatial_learn,Spatial_SC}, shape \cite{DSC_Net,DEC,adversarial_SC,cosineEmbed}, or appearance \cite{video_face_track,visual_feature_clustering} features might not produce satisfactory results. 

The goal of this work is to increase the discriminative capability of spatio-temporal latent representations. Traditional subspace clustering techniques are trained based either on appearance  \cite{DSC_Net,cosineEmbed,adversarial_SC} or location information, generally in the form of bounding boxes \cite{spatial_learn}. Instead, in this work, we propose a novel approach that learns location and shape information jointly using a convolutional and fully connected autoencoder, which we call Deep Heterogeneous Autoencoder (DHAE). To learn a latent representation that leverages motion and appearance information in an unsupervised manner, we employ a multi-task loss function with task-dependent uncertainties \cite{multi_task_learning}. Finally, we use constrained clustering techniques on the latent space to improve the robustness of spatio-temporal data association. In summary, we provide four main contributions:
1) We propose a novel unsupervised mechanism based on task-dependent uncertainties that learns to generate spatially and temporally distinctive latent features based on heterogeneous inputs; 2) We propose a new data partitioning algorithm that uses constrained clustering strategies to associate object detections over multiple frames with their corresponding temporal identifiers; 3) We evaluate our model on two synthetic and two real-world datasets that include most of the challenges commonly observed in MOTS problems, such as pose and appearance variations.

\begin{figure*}[t]
\centering
\includegraphics[width=0.9\textwidth]{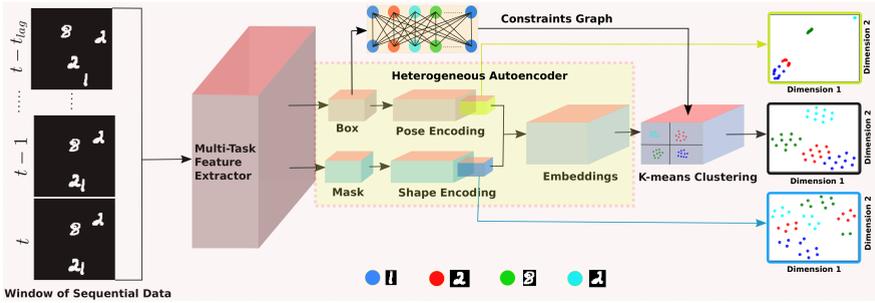}
\begin{small}
\caption {Proposed subspace clustering framework. The Multi-task Feature Extractor detects the bounding boxes and segmentation masks of multiple objects within a window of sequential data. The Deep Heterogeneous Autoencoder then uses these features to generate joint embedded representations of the objects. These embeddings are then clustered into target trajectories using constrained kmeans.} 
\label{fig:model}
\end{small}
\end{figure*}

\section{Related Work}
Subspace clustering is an unsupervised learning technique in which data points are mapped to lower dimensionality subspaces where it is easier to make inferences about the relationships among different data points. Existing clustering methods employ two common strategies: i) extract low-dimensional discriminative features, ii) apply a robust clustering approach to partition the subspaces. Earlier approaches employed methods based on factorization strategies \cite{LRR,elhamifar2013sparse,vidal2011subspace} or kernels \cite{semi_supervised_person_identi,kmeans1,ranzato2007efficient,Mean-Shift-Cheng} to separate the data points into their respective subspaces. 
More recent methods employ convolutional neural networks \cite{DEC,DSC_Net,visual_feature_clustering}, or generative adversarial networks \cite{clusterGAN,adversarial_SC}, oftentimes in conjunction with self-expressive layers \cite{DSC_Net,adversarial_SC,zhang2018scalable}. The autoencoder-based DSC-Net \cite{DSC_Net} uses fully connected layers to learn an affinity matrix that enhances the discriminative property of the embeddings. Some autoencoder-based techniques consider both subspace reconstruction error and cluster assignment error for better sample distribution \cite{yev_autoencoder,DEC}. Although existing approaches may be able to find discriminative features to cluster static data, these features are not sufficiently distinctive to identify the subspaces corresponding to multiple objects in sequential data. 

Clustering multi-object sequential data is an under-explored problem, particularly in real-world applications.  While existing temporal clustering methods, such as ordered subspace clustering (OSC) \cite{OSC} consider sequential data, they focus on clustering entire video frames, without taking into consideration the spatial aspect of the problem, which must be addressed when it is necessary to distinguish multiple objects in a video segment. Few methods \cite{video_face_track,facr_cluster_online_video} address the problem of clustering objects over video sequences, which must take into account the fact that object features may change over time \cite{mozhdehi2018,mozhdehideep3,mozhdehideep4}.  
Our approach addresses this challenge using a simple yet effective unsupervised learning framework.
\section{Subspace Clustering for Sequential Data}

As Fig. \ref{fig:model} illustrates, our spatio-temporal clustering framework extracts features of interest from the objects in each video frame using a multi-task feature extractor module. \begin{minipage}[t]{0.58\textwidth}Then, our proposed DHAE generates a discriminative latent representation of the pose and appearance of each object based on these features. To establish temporal coherence among targets, we adopt a graph-based method \cite{GMCP} to preclude the association of points that violate a set of constraints that are known to hold and enforce the association of points having common temporal identifiers within a temporal window. Finally, we use the constrained kmeans algorithm \cite{cop-kmeans} to determine the labels of the targets by minimizing the dissimilarity of their latent representations while satisfying the association constraints. Alg. \ref{alg:clustering} summarizes the steps of the proposed method, which are described in detail below.
\end{minipage}
\hfill
\begin{minipage}[t]{0.4\textwidth}
\vspace{-1.5em}
\begin{algorithm}[H]
\caption{Subspace clustering} 
\footnotesize
\label{alg:clustering}
     \begin{algorithmic}[1]
         \Require{Set of video frames $\{I^t\}_{t=1}^T$}
         \Ensure{Subspace clusters $\mathcal{C}_K$}
         \Repeat
          \State {$\mathcal{W}^t = \mathtt{MTFE}({\{I^{t'}}\left|t' \in \mathcal{T}^{t} \right.\})$}
          \State {$\mathcal{Z}^t = \mathtt{DHAE}(\mathcal{W}^t)$}
          \State {Compute $\mathcal{G}^{t}$ using Eqs. (\ref{eq:V_w})-(\ref{eq:cl_ml})}
          \State {$\mathcal{C}_K = \emptyset$} 
           \State {$\mathcal{C}_t = \mathtt{kmeans}(\mathcal{Z}^t,\mathcal{G}^{t})$}
           \For {$Q \in \mathcal{C}_t$}
            \State {$\bar{\tau}=\nicefrac{1}{\left |Q \right|}{\sum_{d_i\in Q} \left(c_i\right)}$}
            \If {$\bar{\tau}>\lambda$}
             \State {$\mathcal{C}_K = \mathcal{C}_K \cup \{Q\}$} 
            \EndIf
           \EndFor
        \Until end of the video sequence
   \end{algorithmic}
\normalsize
\end{algorithm}
\end{minipage}
\subsection{Multi-Task Feature Extractor}
The multi-task feature extractor (MTFE) module is responsible for generating segmentation masks and bounding boxes of objects of interest in each video frame. This task is independent of the proposed temporal clustering mechanism and can be performed by any supervised or unsupervised segmentation method such as \cite{DBLP:journals/corr/HeGDG17,kim2019mumford}. More specifically, let $x_{b,i}^t\in \mathbb{R}^{N}$ be the detected bounding box of the $i$-th target observed at time $t$, $x_{m,i}^t \in \mathbb{R}^{M\times M\times D}$ be a segmentation mask representing the appearance of that object, and $c_i^t \in [0,1]$ be the corresponding detection confidence. The MTFE takes as input a video frame $I^t$ and generates the set
\begin{equation}
\mathcal{X}^t=\left\{[x_{m,i}^{t},x_{b,i}^{t}, c_i^t] \right\}_{i=1}^{\mathcal{O}^t}, \label{eq:Xt}
\end{equation}
where $\mathcal{O}^{t}$ is the number of unique objects at time $t$.  The bounding box $x_{b,i}^t$ is represented by the coordinates of its centroid and its dimensions, hence $N=4$. Regarding  appearance representation, we propose two closely related models. In the \emph{shape} model, the number of channels of the mask $D=1$ and $x_{m,i}^t$ corresponds to the binary segmentation mask of the object. In the \emph{appearance} model, $D=3$ and $x_{m,i}^t$ is given by the binary segmentation mask multiplied by the corresponding RGB contents of the image.

\subsection{Deep Heterogeneous Autoencoder} \label{sec:pagestyle}
Clustering methods that resort only to object appearance information do not perform well when multiple objects are observed simultaneously in a sequence of video frames. As the number of objects of a certain category (e.g., pedestrians) observed in a given frame increases, the average appearance difference among them becomes increasingly lower. At the same time, as the duration of the temporal segment increases, so does the variability in the appearance of any given target. Hence, to allow for sufficient temporal appearance variability while preventing incorrect associations among temporally proximal observations, we incorporate location information into the latent feature representation. 

\begin{wrapfigure}{r}{0.47\textwidth}
\vspace{-1.1em}
\centering
\includegraphics[width=\linewidth]{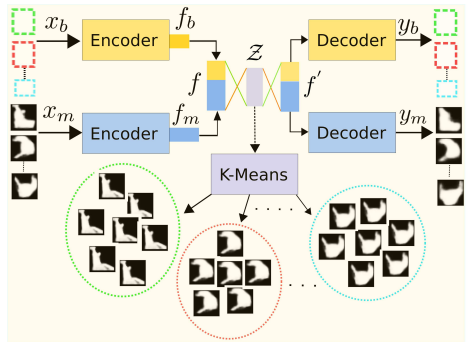}
\vspace{-1em}
\caption {Proposed Deep Heterogeneous Autoencoder (DHAE) model architecture, which jointly learns shape and location information for sequential data clustering.}
\label{fig:CAE}
\vspace{-1.5em}
\end{wrapfigure}

Fig. \ref{fig:CAE} shows our proposed DHAE architecture. 
To combine shape and location information, we design a network consisting of three parts: i) a pair of encoders that take 
as input the $N$-dimensional location vector\footnote{To simplify the notation, we henceforth drop the subscript $i$ and the superscript $t$.} $x_{b}$ and the $M\times M\times D$ mask $x_{m}$, ii) an uncertainty-aware module based on self-expressive layers
\cite{DSC_Net} to reconstruct the concatenated feature $f'$ and learn the latent feature $\mathcal{Z}$, iii) a pair of decoders to reconstruct the bounding box $y_{b}$ and the mask $y_{m}$. The DHAE takes the extracted set of shapes and locations $\mathcal{X}^t$, which are generated by the MTFE, and reconstructs them by minimizing the combined reconstruction loss. To incorporate the location features $x_b$ into our model, we employ a fully 
connected auto-encoder (AE) with $N$ inputs, which is represented by the yellow boxes in Fig. \ref{fig:CAE}. The
corresponding encoded feature vector is $f_{b}=h_{b}(x_b)$, where $h_{b}:\mathbb{R}^N\rightarrow\mathbb{R}^{F}$ is the encoding function. The shape information $x_m$ is encoded by a convolutional auto-encoder (CAE) with an input size of ${M\times M\times D}$, which is represented by the blue boxes in Fig. \ref{fig:CAE}. Let $f_{m}=h_{m}(x_m)$ be the latent feature of the CAE, where $h_{m}:\mathbb{R}^{M\times M\times D}\rightarrow \mathbb{R}^{F}$ is the encoding function. A function $h_{a}:\mathbb{R}^{2F}\rightarrow\mathbb{R}^{F}$ takes the concatenated feature vector $f=[f_{m},f_{b}]$ and converts it into the latent representation $\mathcal{Z} \in \mathbb{R}^{F}$.
A function $h_{a}':\mathbb{R}^{F}\rightarrow\mathbb{R}^{2F}$ then takes the latent representation $\mathcal{Z}$ and produces a new feature $f^{'}=[f^{'}_{m},f^{'}_{b}]$, which combines both shape and location information. We then process the two components of the feature vector separately using the corresponding decoders.
That is, the outputs produced by the decoding function $h^{'}_{m}:\mathbb{R}^{F}\rightarrow\mathbb{R}^{M\times M\times D}$ and $h^{'}_{b}:\mathbb{R}^{F}\rightarrow\mathbb{R}^{N}$are $y_{m}\in\mathbb{R}^{M\times M\times D}$ and $y_{b}\in\mathbb{R}^{N}$.


\subsubsection{Multi-task Likelihood}
We train our DHAE using a multi-task loss function using maximum log-likelihoods with task-dependent uncertainties. Let $f^{W}(x)=[f_{m}^{W}(x),f_{b}^{W}(x)]$ be the output of the DHAE with weights $W$, input vector $x=[x_{m},x_{b}]$, and predicted output $y=[y_{m},y_{b}]$. The likelihood for the regression task is given by 
\begin{equation}
\begin{aligned}p(y_{m},y_{b}|f^{W}(x),\sigma_m,\sigma_b)=p_{m}(y_{m}|f_{m}^{W}(x),\sigma_m)\cdot p_{b}(y_{b}|f_{b}^{W}(x),\sigma_b),\label{eq:likelihhod}\end{aligned}
\end{equation}
where $y_{b}$ and $y_{m}$ are normally distributed with means $f_{b}^{W}(x)$,  $f_{m}^{W}(x)$, and variances $\sigma_{b}$, $\sigma_{m}$, respectively. Thus, the cost function is given by 
\begin{multline}
\mathcal{L}(W,\sigma_{m},\sigma_{b})=-\log p(y_{m},y_{b}|f^{W}(x), \sigma_b, \sigma_m)\\
\propto\frac{1}{2\sigma_{b}^{2}}\left\Vert y_{b}-f_{b}^{W}(x)\right\Vert ^{2}+\frac{1}{2\sigma_{m}^{2}}\left\Vert y_{m}-f_{m}^{W}(x)\right\Vert ^{2} +\log\sigma_{m}+\log\sigma_{b},
\label{eq:loss_func}
\end{multline}
where $W$, $\sigma_m$, and $\sigma_b$ are trainable parameters. Unlike traditional multi-task learning approaches \cite{multitask-segment,multi_task_learning}, which focus on weighing the contributions of several \emph{homogeneous outputs}, our method weighs the contribution of multiple \emph{heterogeneous input} features. Thus, we learn the relative weights of each loss function term based on the uncertainty of distinct 
features.

\subsubsection{Network Implementation and Training Details}
\label{sec:typestyle}
The AE branch of our DHAE has an input size of $N=4$ and one fully connected layer of size 128. The CAE branch uses 5 convolutional layers with kernel size $3\times 3$, ReLU activations, and a stride of $2\times 2$ for downsampling and upsampling \cite{dumoulin2018guide}. The size of the input layer is $M=128$ and the subsequent layers have half the size of the previous layer. The number of convolutional channels in each layer is 16, 16, 32, 32, and 64 (the decoder mirrors the structure of the encoder). 
The functions $h_a(\cdot)$ and $h_a'(\cdot)$ are implemented using fully connected layers of size $F=128$. We train the network using stochastic gradient descent with ADADELTA \cite{zeiler2012adadelta} learning rate adaptation.
To account for the dimensionality of the feature vectors, we initialize $\log\sigma_b^2=\nicefrac{1}{N}$ and $\log\sigma_m^2=\nicefrac{1}{\left (M^2 \right )}$ 
. The network weights $W$ are initialized using the Glorot method \cite{glorot2010understanding}. At inference time, the segmentation masks are isotropically scaled such that the largest dimension of the mask is $M$. The image is then centered along the smallest dimension and zero-padded.

\begin{figure*}[t]
\centering
\includegraphics[width=.95\linewidth]{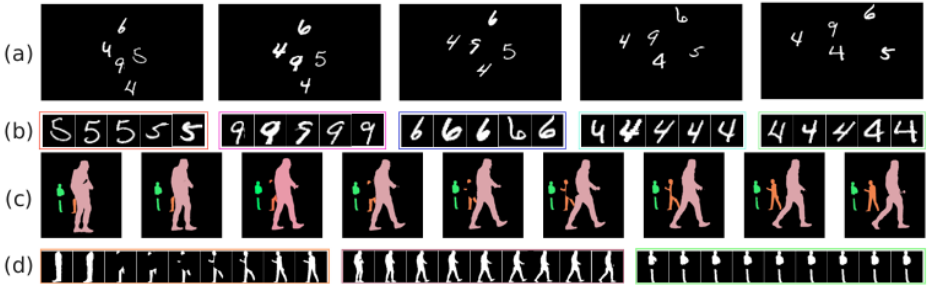}
\begin{small}
\vspace{-1em}
\caption {Sequence of frames for (a) MNIST-MOT with $|O|=5$ targets per frame and (c) MOTSChallenge (cropped for three pedestrians) sequences. (b) and (d) show the corresponding subspaces generated by our method.} 
\label{fig:subspace_example}
\end{small}
\end{figure*}

\subsection{Sequential Data Constraints}
\label{sub:sequential}
In spatio-temporal subspace clustering, constraints based on prior knowledge regarding the sequential data may reduce association mistakes by imposing penalties on unlikely pairwise matching. We use an undirected graph $\mathcal{G}^{t}$ to encode constraints among pairs of detections in the temporal window $\mathcal{T}^t$. This graph determines which pairs of detections cannot belong to the same
cluster. It also enforces the association of detections that were assigned the same temporal identifier in previous temporal windows. 

\subsubsection{Constraints Graph Formulation}
\label{Constraints Graph Formulation}
To incorporate prior knowledge regarding object correspondences in a temporal window, we construct the graph $\mathcal{G}^{t}=(V^t,E^t)$ using \emph{cannot link} and \emph{must link} constraints. The vertices of the graph correspond to the set of segmentation masks in all the frames in the window $\mathcal{T}^t$, i.e.,
\begin{equation}
V^t=\left\{ x_{m,i}^{t}\left|t\in \mathcal{T}^t, i \in \left\{1, \ldots, \mathcal{O}^t \right\} \right.\right\}.
\label{eq:V_w}
\end{equation}
The set of edges consists of pairs of nodes $v_i$ and $v_j$ that meet the spatial and temporal restrictions imposed by the \emph{cannot link function} $f_{cl}(\cdot)$ and \emph{must link function} $f_{ml}(\cdot)$, i.e.,
\begin{equation}
\begin{aligned}
E^t_{cl} &= \left\{(v_i,v_j)|v_i \in V^t, v_j \in V^t, f_{cl}=1 \right\}, & 
E^t_{ml} &= \left\{(v_i,v_j)|v_i \in V^t, v_j \in V^t, f_{ml}=1 \right\}.
\label{eq:E_w}
\end{aligned}
\end{equation}
The function $f_{cl}$ prevents the association of vertices from the same frame or vertices in close temporal proximity whose segmentation masks do not overlap. Conversely, $f_{ml}$ enforces the association of detections with common temporal identifiers. That is, if a detection has been assigned an identifier at a previous temporal window, it is only allowed to be associated with detections that have the same identifier or that have not been assigned one, i.e.,
\vspace{-1.5em}
\begin{equation}
\begin{minipage}[H]{0.5\linewidth}
$$\begin{aligned}
f_{cl}(v_i,v_j)=\begin{cases}
 1&  \text{ if } t_i=t_j \text{ or } \gamma_i \neq \gamma_j\\ 
 1& \text{ if } \mathtt{iou}(v_i,v_j)=0, \ t_i - t_j\leqslant \tau \\
 0 & \text{otherwise}
\end{cases},
\end{aligned} $$

\end{minipage} 
\hspace{10pt} 
\begin{minipage}[H]{0.5\linewidth}
$$\begin{aligned}
f_{ml}(v_i,v_j)=\begin{cases}
 1&  \text{ if } l_i = l_j,\ \gamma_i= \gamma_j, \\
 &t_i \neq t_j, \ l \neq \emptyset \\ 
 0& \text{otherwise}
\end{cases},
\label{eq:cl_ml} 
\end{aligned}$$
\end{minipage}
\end{equation}
where $t_i$, $t_j$ are the timestamps for the detections corresponding to $v_i$ and $v_j$, the function $\mathtt{iou}(\cdot)$ computes their mask intersection over union, $\gamma_i$, $\gamma_j$ are the corresponding object classes, and $l_i$, $l_j$ are the initialized cluster identities corresponding to nodes $v_i$ and $v_j$. 

\subsubsection{Modified Constrained Kmeans}
\label{constrained k-means}
In the constrained kmeans algorithm \cite{cop-kmeans}, a node $v_i$ may be assigned to the same cluster as $v_j$ only if the edge $(v_i,v_j) \notin E^t_{cl}$ and the initialized detections maintain the same subspace clustering when $(v_i,v_j) \in E^t_{ml}$. We cluster objects that do not satisfy the \emph{cannot link} constraints and whose tracklet identities are not yet initialized by minimizing the distance between their corresponding latent features $z_i^t$ and the cluster centroids $z_k$.
 
 The dimensionality of the subspace corresponding to window $\mathcal{T}^t$ is given by the total number of objects observed during that period, which is unknown. We propose a novel mechanism to determine the number of clusters $|\mathcal{K}|$ that leverages our temporal clustering constraints. Our approach consists of initially setting the number of clusters to the maximum number of targets observed in a single frame within the window $\mathcal{T}^t$, i.e., $|\mathcal{K}|=\operatorname*{max}_{t' \in \mathcal{T}^t} \mathcal{O}^{t'}.$
The centroids of the clusters are then initialized using the detections in $\mathcal{O}^{t'}$. Then, if the clustering constraints cannot be satisfied, the number of clusters is adjusted accordingly. That is, if a detection does not satisfy the constraints in Eq. \ref{eq:cl_ml}, it is considered a tentative new target, a new cluster is created, and as new detections join that cluster in subsequent frames, a new tracklet is generated. If no subsequent detections join the cluster within a $t_{lag}$ interval, the new detection is considered a mistake and the corresponding cluster is discarded.

\subsection{Spatio-temporal Clustering}
To assign unique temporal identifiers to multiple objects in a video sequence, at each time instant $t$, we cluster the observations present in the frames within the window $\mathcal{T}^t=\{t-t_{lag}, t-t_{lag+1}, \ldots, t\}$ (Fig. \ref{fig:subspace_example}). Since the window is computed at each frame, there is an overlap of $t_{lag}-1$ frames between subsequent windows, which ensures that most of the data points used to form the subspace cluster in each window of a video sequence are shared. As shown in Alg. \ref{alg:clustering}, the input to our spatio-temporal clustering algorithm is the set of frames $\{I^t\}_{t=1}^T$ where $T$ is the number of frames in the video. Our algorithm applies the MTFE to each video frame in the window $\mathcal{T}^t$ to construct the set $\mathcal{W}^t=\left\{\mathcal{X}^{t'}|t'\in \mathcal{T}^t\right\}$, where $\mathcal{X}^{t'}$ is 
  the output of the MTFE for frame $I^{t'}$. The algorithm then clusters the detections within each window using the embeddings $\mathcal{Z}^t=\left\{z_i^{t'} \left| t' \in \mathcal{T}^{t}, i\in \{1, \ldots, \mathcal{O}^{t'} \} \right. \right\}$ generated by the DHAE. 
Each call to the $\mathtt{kmeans(\cdot)}$ algorithm produces a set of clusters $\mathcal{C}_t$ whose elements are detections assigned to the same object. 
For each cluster $Q\in \mathcal{C}_t$, we compute its normalized score $\bar{\tau}$ as the ratio $\nicefrac{1}{|Q|}\sum_{d_i\in Q}(c_i)$, where $c_i$ is the confidence score of detection $d_i$. Clusters with a  score higher than a threshold $\lambda$ are included in the cluster set $\mathcal{C}_K$.
\section{Datasets and Experiments}
\label{sec:print}
We evaluate our algorithm on two synthetic and two real-world datasets. The MNIST-MOT and Sprites-MOT \cite{DBLP:tracking_by_animation} synthetic datasets allow us to simulate challenging MOTS scenarios involving pose, scale, and shape variations. In addition, their bounding box and segmentation masks are readily available. Then, we use the recently published MOTS \cite{MOTS} benchmark, which includes video sequences from the MOTChallenge \cite{MOT16} and the KITTI \cite{KITTI} datasets annotated with segmentation masks,  
to evaluate our model in real-world videos. 

Since traditional clustering measures \cite{adversarial_SC,NMI1,NMI2} require each observation to be mapped to exactly one cluster, they are not suitable for real-world scenarios where incorrect detections may occur. Hence, we adopt the popular CLEAR-MOT \cite{Stiefelhagen:CLEAR_MOT} tracking performance assessment measures: Multi-object Tracking Accuracy (MOTA), Fragmentation (Frag), Identity Switches (IDs), Mostly Tracked (MT), and Mostly Lost (ML) targets \cite{track_quality_measures}. Finally, we employ the MOTS \cite{MOTS} evaluation measures to quantify the effectiveness of our algorithm in maintaining the temporal consistency of target identities in real-world video sequences. 
\vfill
\subsection{Synthetic Datasets}
We generate synthetic MNIST-MOT and Sprites-MOT sequences using the procedure described in \cite{DBLP:tracking_by_animation}, which includes most of the common challenges observed in MOT problems. For the MNIST-MOT dataset, we generate 9 digit classes and for Sprites-MOT we generate 4 geometric shapes. For both datasets, the object density is $|O|=3$, the target birth probability is $0.5$, the size of each object is $28 \times 28$ at each frame of size $128 \times 128$, and the average target velocity is $5.3$ pixels per frame. We generate 20 video sequences with 500 frames for each dataset. In each sequence, the set of initial objects (digits or sprites) is chosen randomly at the first frame. In subsequent frames, the objects move in random directions. 

To train the DHAE, we extract the bounding box and shape mask of each object from the synthetic video frames using a separate set of training sequences. Due to the lack of availability of methods that perform clustering based on location and shape features, we select one state-of-the-art MOT method for performance comparison \cite{DBLP:tracking_by_animation}. 
We relax the IoU constraint by imposing a limit on the Euclidean distance between embeddings since the target displacement among consecutive frames may be relatively large with respect to the size of the targets due to the low resolution of the frames.

\begin{table*}[ht]
\centering
\caption{Performance evaluation on MNIST-MOT and Sprites-MOT with $t_{lag} = 3$, $|O| = 3$. 
}
\label{tab:MNIST-MOT-Sprites-MOT-Baseline}
\setlength{\tabcolsep}{1.8pt} 
\footnotesize
\begin{tabular}{l|ccccccc|ccccccc}
\hline
\multirow{2}{*}{Method}    & \multicolumn{7}{c|}{MNIST-MOT} & \multicolumn{7}{c}{Sprites-MOT} \\
\cline{2-15}
 & $\uparrow$IDF1 & $\uparrow$MT &$\downarrow$ML &$\downarrow$FN &$\downarrow$IDs &$\downarrow$Frag &$\uparrow$MOTA   & $\uparrow$IDF1   &$\uparrow$MT &$\downarrow$ML &$\downarrow$FN &$\downarrow$IDs &$\downarrow$Frag &$\uparrow$MOTA   \\
\hline
shape embed        &89.1 &944 &0  &304 &5 &132 &98.6      &86.7 &906  &4 &853 &13  &275 &96.1 \\
loc embed           &87.7 &905 &0 &708 &61 &331 &96.5      &88.2 &920 &0 &689 &56 &347 &96.6    \\  
loc+$\mathcal{G}^t$  &86.3 &977 &0  &0 &72 &0 &99.7      &85.6 &983 &0 &4 &94 &0 &99.6     \\
loc+shape           &89.6 &934 &0  &444 &3 &189 &98.0     &88.9 &912 &0 &734 &\textbf{8} &314 &96.6       \\
\textbf{loc+shape+$\mathcal{G}^t$}  &\textbf{100.0} &977 &\textbf{0}  &\textbf{0} &\textbf{0} &\textbf{0} &\textbf{100.0}     &\textbf{99.5} &983 &\textbf{0} &\textbf{45} &17 &\textbf{0} &\textbf{99.7}    \\
TBA \cite{DBLP:tracking_by_animation}   &99.6 &\textbf{978} &0  &49 &22 &7 &99.5     &99.2 &\textbf{985} &1 &80 &30 &22 &99.2     \\
\hline
\end{tabular}
\normalsize
\end{table*}

Table \ref{tab:MNIST-MOT-Sprites-MOT-Baseline} summarizes the performance of our method on the MNIST-MOT and Sprites-MOT datasets according to the evaluation procedure described in \cite{Stiefelhagen:CLEAR_MOT,DBLP:tracking_by_animation, track_quality_measures}. Although location features play a critical role in clustering multiple moving targets, shape features also contribute significantly to the performance of our approach. As the t-SNE visualization in Fig. \ref{fig:model} illustrates, as the digits 1 and 2 approach each other, their embeddings remain separable.
The shape-only model (\emph{shape embed}) outperforms the location-only model (\emph{loc embed}) on some of the evaluation criteria because the shapes remain unchanged until they leave the scene. This effect is more pronounced on the MNIST-MOT dataset because the appearance of the characters is more distinctive than the shapes in Sprites-MOT.
Although the incorporation of the constraints graph into the location-only model (\emph{loc+}$\mathcal{G}^t$) leads to slightly better performance on some evaluation measures than the joint embedding  (\emph{loc+shape}), the overall method (\emph{loc+shape+}$\mathcal{G}^t$) achieves near-perfect results.



\subsection{MOTS Dataset}
The MOTSChallenge dataset consists of four fully annotated videos of crowded scenes and the KITTI MOTS dataset consists of 21 videos acquired from a moving vehicle. Both datasets contain objects that show substantial scale and shape variations over time. We use the segmentation masks and the corresponding RGB content from 12 KITTI MOTS sequences to train the DHAE and use the remaining sequences for testing. 

\begin{table}[t]
\begin{minipage}[t]{0.48\textwidth}
\caption{Evaluation of person and car tracking on the KITTI MOTS validation set.}
\label{tab:MOTS-KITTI}
\centering
\setlength{\tabcolsep}{2.8pt} 
\footnotesize
\begin{tabular}{@{}lllllll@{}}
\hline
\multicolumn{1}{c}{\multirow{2}{*}{Method}} & \multicolumn{2}{l}{$\uparrow$sMOTSA} & \multicolumn{2}{l}{$\uparrow$MOTSA} & \multicolumn{2}{l}{$\uparrow$MOTSP} \\
\multicolumn{1}{c}{}                        & car          & ped         & car         & ped         & car         & ped         \\ 
\hline
$\text{$\text{loc+shape+$\mathcal{G}_t$}$}$                                  &80.4               &55.5            &89.5             &69.7              &90.3             &82.8            \\
\textbf{$\text{$\text{loc+app+$\mathcal{G}_t$}$}$}                                  &\textbf{80.8}               &\textbf{58.3}            &\textbf{89.9}            &\textbf{72.5}              &\textbf{90.3}             &\textbf{82.8}             \\
\hline
$\text{EagerMOT \cite{EagerMOT}}$                                 &74.5              &58.1             &-            &-             &-             &-             \\
$\text{GMPHD \cite{GMPHD2020online}}$                                 &76.9              &48.8             &-            &-             &87.1             &76.4             \\
$\text{MOTSFusion  \cite{MOTSFusion}}$                                 &77.5              &49.9             &89.2            &66.6             &-             &-             \\
$\text{MOTSNet  \cite{MOTSNet}}$                                 &78.1              &54.6             &87.2             &69.3             &89.6             &79.7             \\
$\text{TR-CNN  \cite{MOTS}}$                                 &76.2              &46.8             &87.8             &65.1             &87.2             &75.7             \\
\hline
\end{tabular}
\normalsize
\end{minipage} %
\hspace{0.5em}
\begin{minipage}[t]{0.47\textwidth}
\caption{Evaluation of person tracking on the MOTSChallenge training set.}
\label{tab:MOTSChallenge}
\centering
\setlength{\tabcolsep}{1.8pt} 
\footnotesize
\begin{tabular}{lcccccl}
\hline
Method     &$\uparrow$sMOTSA  &$\uparrow$MOTSA  &$\uparrow$MOTSP &$\uparrow$MT \\ \hline

 $\text{loc+shape}$ &51.3 &60.1 &86.3 &21.5
  \\
  $\text{loc+shape+$\mathcal{G}_t$}$ &65.3 &76.3 &86.3 &53.1
  \\
$\text{loc+app}$ &56.1 &65.5 &\textbf{86.4} &26.3
  \\
 $\text{loc+app+$\mathcal{G}_t$}$ &65.5 &76.5 &86.3 &\textbf{53.1}
 \\
 $\textbf{\text{GMPHD \cite{GMPHD2020online}}}$ &\textbf{65.8} &\textbf{77.1} &86.1 &-
 \\
 $\text{PointTrack \cite{point_track_eccv2020}}$ &58.1 &70.6 &- &-
 \\ 
 $\text{MOTSNet \cite{MOTSNet}}$ &56.8 &69.4 &82.7 &-
 \\ 
$\text{TR-CNN \cite{MOTS}}$ &52.7 &66.9 &80.2 &-
 \\ 
\hline
\end{tabular}
\normalsize
\end{minipage} %
\end{table}

\begin{figure*}[t]
\centering
\begin{small}
\includegraphics[width=.99\linewidth]{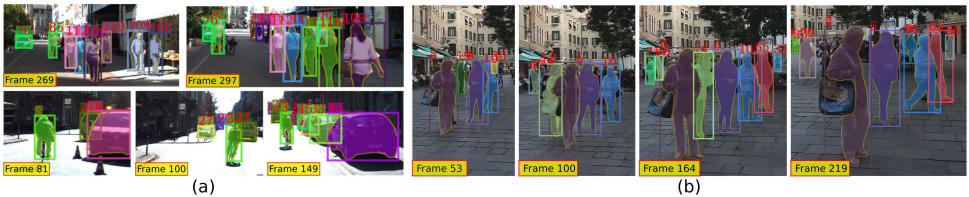}
\caption {Qualitative results on the  (a) KITTI MOTS and (b) MOTSChallenge dataset.}
\label{fig:quali_kitti_mot17}
\end{small}
\end{figure*}

In our evaluation, we use the publicly available instance segmentation masks and bounding boxes \cite{MOTS} from the benchmark validation set. 
We compare the performance of our method against the state-of-the-art approaches presented in \cite{EagerMOT,MOTSFusion,MOTS,MOTSNet,GMPHD2020online,point_track_eccv2020}. 
Table \ref{tab:MOTS-KITTI} shows that, without resorting to sophisticated mechanisms for target re-identification, trajectory interpolation, or entry/exit detection, our method outperforms all the baseline methods in the KITTI MOTS dataset even if only the binary segmentation masks are used in the joint embeddings (\textit{loc+shape+}$\mathcal{G}_t$). Including the RGB information (\textit{loc+app+}$\mathcal{G}_t$) leads to further performance gains, particularly for the pedestrian class. As Table \ref{tab:MOTSChallenge} indicates, in the MOTSChallenge sequences, our joint embeddings alone (\textit{loc+app}) perform on par with \cite{MOTSNet,MOTS,point_track_eccv2020}, and the incorporation of the constraints graph leads to results comparable to \cite{GMPHD2020online} using the same set of detections (whereas \cite{point_track_eccv2020} and \cite{MOTSNet} use privately refined detections).
Figure \ref{fig:quali_kitti_mot17} illustrates some of the results generated by our method. Both datasets are comprised of crowded scenes with significant amounts of temporary partial and full occlusions, particularly among pedestrians. Unlike \cite{GMPHD2020online}, our method does not incorporate occlusion reasoning or motion modelling techniques, which contribute significantly to the performance.

\begin{wrapfigure}{r}{0.38\textwidth}
\vspace{-1em}
\centering
\includegraphics[width=.95\linewidth]{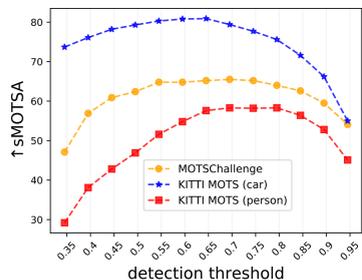}
\vspace{-1em}
\caption {MOTS performance as a function of MTFE detection score threshold.} 
\label{fig:det_score_vs_sMOTS}
\vspace{-0.3em}
\end{wrapfigure}
To assess the impact of detection noise on the performance of our method, we evaluate the sMOTSA measure as a function of the minimum confidence score for a detection to be considered valid. Fig. \ref{fig:det_score_vs_sMOTS} shows that performance increases until the detection threshold reaches approximately $0.65$ and after $0.80$ it starts to decrease again. 
In the experiments discussed above, we use a detection threshold of $0.70$ for all the datasets. Further performance improvements would be achieved with dataset-specific thresholds. For our ablation studies, we use ground truth annotations instead 
to evaluate each step of the method independently from the performance of the underlying detector.

\subsection{Ablation Study}

Table \ref{tab:ablation-MOTS-noiseless} shows the impact of the uncertainty-aware multitask learning loss of Eq. \ref{eq:loss_func}, of the constraints graph $\mathcal{G}^t$, and of estimating the number of clusters $|\mathcal{K}|$ using the method described in Section \ref{constrained k-means} for three different window sizes $t_{lag}$. The table demonstrates the positive impact of multi-task learning using task uncertainties instead of assigning equal weights in Eq. \ref{eq:loss_func}. We observe that the constraints graph leads to consistent and substantial improvements in all the evaluation criteria. We also see that higher values of $t_{lag}$ lead to an increase in the MT measure but also to an increased number of fragmentations. Since we do not model target motion, the overlap among detections reduces as $t_{lag}$ increases, leading to violations of the must-link constraints. 
As a result, the optimal sMOTSA score is obtained with $t_{lag}=3$. Finally, estimating $|\mathcal{K}|$ from the video segments does not degrade the performance of our algorithm.  In summary, Table \ref{tab:ablation-MOTS-noiseless} shows that MTL and $\mathcal{G}^t$ lead to improvements in the sMOTSA measure of $10.8\%$ and $15.5\%$ in the MOTSChallenge,  $8.2\%$ and $14.1\%$ for the car category in the KITTI MOTS, and $11.1\%$ and $26.2\%$ for the person class, even when $|\mathcal{K}|$ is unknown.


\begin{table}[ht]
\centering
\caption{Ablation study for different components of our method. We evaluate the impact on tracking based on MOTS \cite{MOTS} oracle performance of:  window size $t_{lag}$, constraints graph $\mathcal{G}^t$, multi-task learning (MTL), and number of subspaces $|\mathcal{K}|$ as a prior (\xmark) or estimated (\checkmark).}
\label{tab:ablation-MOTS-noiseless}
\setlength\tabcolsep{1.7pt} 
\footnotesize
\begin{tabular}{cccccccccccccccc}
\hline
&       &          &                     & \multicolumn{4}{c}{MOTSChallenge}                          & \multicolumn{4}{c}{KITTI MOTS (car)} & \multicolumn{4}{c}{KITTI MOTS (person)}\\
$t_{lag}$  &$\mathcal{G}^t$ & MTL  & $|\mathcal{K}|$ &$\uparrow$sMOTSA    &$\uparrow$MT    &$\downarrow$IDs  &$\downarrow$Frag    &$\uparrow$sMOTSA    &$\uparrow$MT    &$\downarrow$IDs  &$\downarrow$Frag   &$\uparrow$sMOTSA    &$\uparrow$MT    &$\downarrow$IDs  &$\downarrow$Frag     \\
\hline
\multirow{5}{*}{3} &\xmark     &\xmark     &\checkmark      &74.0 &43.9 &962 &1835   &77.5 &63.6 &217 &479  &62.7 &57.4 &255 &318   \\
&\xmark     &\checkmark &\checkmark      &82.6 &62.3 &660 &1416   &84.4 &68.2 &115 &348  &70.5 &54.4 &138 &258   \\
&\xmark     &\checkmark &\xmark  &84.0 &68.9 &653 &1367  &84.7 &70.9 &113 &343   &71.7 &52.9 &138 &248  \\
&\checkmark &\checkmark &\xmark     &97.5 &\textbf{100} &636 &639  &98.1 &98.7 &127 &127   &95.6 &98.7 &119 &\textbf{115}  \\
&\checkmark &\checkmark &\checkmark  &\textbf{97.8} &99.6 &548 &\textbf{548}  &\textbf{98.3} &\textbf{98.7} &119 &\textbf{121}   &\textbf{95.6} &98.5 &124 &124   \\
\hline

\multirow{5}{*}{5} &\xmark     &\xmark     &\checkmark &66.4 &31.1 &939 &2000   &75.5 &58.9 &191 &474   &53.0 &52.9 &226 &282   \\
&\xmark     &\checkmark &\checkmark      &78.9 &50.9 &606 &1476   &76.7 &56.3 &97 &408  &63.0 &32.4 &100 &238   \\
&\xmark     &\checkmark &\xmark  &80.9 &58.3 &610 &1422  &77.4 &55.6 &94 &409   &63.2 &36.8 &92 &254   \\
&\checkmark &\checkmark &\xmark     &97.2 &100 &692 &733  &97.7 &98.7 &160 &171   &94.1 &100 &186 &199   \\
&\checkmark &\checkmark &\checkmark  &97.7 &100 &581 &589  &97.8 &98.7 &162 &162   &94.6 &\textbf{100} &180 &181   \\
\hline

\multirow{5}{*}{8} &\xmark     &\xmark     &\checkmark      &56.9 &19.7 &897 &1800   &71.4 &50.3 &218 &493  &46.0 &38.2 &141 &227   \\
&\xmark     &\checkmark &\checkmark      &71.8 &41.7 &\textbf{447} &1499   &73.0 &47.7 &79 &376  &58.2 &30.9 &83 &246   \\
&\xmark     &\checkmark &\xmark &73.6 &46.9 &528 &1440  &74.8 &47.0 &\textbf{77} &384   &57.5 &35.3 &\textbf{82} &250   \\
&\checkmark &\checkmark &\xmark     &97.0 &99.1 &765 &800  &96.8 &98.0 &223 &236   &93.5 &100 &215 &217   \\    
&\checkmark &\checkmark &\checkmark  &97.5 &100 &657 &662  &97.2 &98.0 &210 &210   &94.1 &100 &194 &196   \\
\hline
\end{tabular}
\normalsize
\end{table}
\section{Conclusions}
\label{sec:illust}
Our proposed method uses task-dependent uncertainties to simultaneously learn the contribution of shape/appearance and location features from multi-object video datasets and further improves clustering performance by imposing simple constraints on acceptable sequential data patterns. Our experimental results show that our approach can accurately cluster multiple objects using embeddings generated by the DHAE. This method can be extended without significant modifications to include additional tasks of interest in similar scenarios such as object motion prediction. In the future, we intend to extend our method with target motion models and more robust entry/exit/occlusion detection techniques so that it can be employed as a robust, standalone data association mechanism to the problems of multiple object tracking \cite{sun2019deep} and video instance segmentation \cite{luiten2019video}.

\bibliography{egbib}
\end{document}